\documentclass[lettersize,journal]{IEEEtran}
\usepackage{amsmath,amsfonts}
\usepackage{algorithmic}
\usepackage{array}
\usepackage[caption=false,font=normalsize,labelfont=sf,textfont=sf]{subfig}
\usepackage{textcomp}
\usepackage{stfloats}
\usepackage{url}
\usepackage{verbatim}
\usepackage{graphicx}
\usepackage{colortbl}

\def\BibTeX{{\rm B\kern-.05em{\sc i\kern-.025em b}\kern-.08em
    T\kern-.1667em\lower.7ex\hbox{E}\kern-.125emX}}
\usepackage{balance}

\usepackage{amsmath}
\usepackage{amsthm}
\usepackage{booktabs}
\usepackage{algorithm}
\usepackage[switch]{lineno}
\usepackage{amssymb}
\usepackage{booktabs}
\usepackage{makecell}
\usepackage{multirow}
\usepackage{cancel}
\usepackage{xcolor}
\usepackage{nicematrix}
\usepackage{listings}
\usepackage[numbers,sort]{natbib}

\newcommand{\blue}[1]{\textcolor{blue}{#1}}

\newcommand\clearrow{\global\let\rowmac\relax}

\newtheorem{proposition}{Property}

\definecolor{Gray}{gray}{0.9}

\begin{document}
\title{Life Regression based Patch Slimming for Vision Transformers}

\author{
    Jiawei Chen,
    Lin Chen,
    Jiang Yang,
    Tianqi Shi,
    Lechao Cheng,
    Zunlei Feng,\\
    Mingli Song~\IEEEmembership{Senior~Member,~IEEE}
\thanks{Jiawei Chen, Lin Chen, Zunlei Feng and Mingli Song are with the College of Computer Science and Technology, Zhejiang University, Hangzhou, China. e-mail: brooksong@zju.edu.cn
}
\thanks{
Lechao Cheng is with Zhejiang Lab. e-mail: chenglc@zhejianglab.com
}
\thanks{
Jiang Yang and Tianqi Shi are with Alibaba Group, Hangzhou, China. e-mail: yangjiang.yj@alibaba-inc.com
}
}

\maketitle

\begin{abstract}
Vision transformers have achieved remarkable success in computer vision tasks by using multi-head self-attention modules to capture long-range dependencies within images. However, the high inference computation cost poses a new challenge. Several methods have been proposed to address this problem, mainly by slimming patches. In the inference stage, these methods classify patches into two classes, one to keep and the other to discard in multiple layers. This approach results in additional computation at every layer where patches are discarded, which hinders inference acceleration. 

In this study, we tackle the patch slimming problem from a different perspective by proposing a life regression module that determines the lifespan of each image patch in one go. During inference, the patch is discarded once the current layer index exceeds its life. Our proposed method avoids additional computation and parameters in multiple layers to enhance inference speed while maintaining competitive performance. Additionally, our approach requires fewer training epochs than other patch slimming methods.
\end{abstract}

\begin{IEEEkeywords}
Vision Transformer, Acceleration, Model Optimization.
\end{IEEEkeywords}

\section{Introduction}
\IEEEPARstart{T}{he} Vision Transformer model architecture has emerged as a leading approach for diverse computer vision tasks, including image classification, object detection, and semantic segmentation, owing to the breakthrough success of the ViT model architecture~\cite{dosovitskiy2020image}. The ViT model divides input images into patches of equal sizes that are then processed as input. Each ViT layer comprises a multi-head self-attention module and an MLP module with residual connections to capture long-range dependencies. Overall, multiple variations of ViT models have been introduced that aim to improve the overall performance, trainability, and data efficiency of the model~\cite{liu2021swin,bao2021beit,jiang2021all,touvron2021training}.

Despite the recent advances in vision transformers, the high computation cost for inference has emerged as a major challenge for deploying such models in real-time scenarios~\cite{touvron2021training}. Compared to CNNs, vision transformers lack important inductive bias, such as translation equivariance, and require a greater number of training/fine-tuning epochs, which can slow down the deployment process~\cite{touvron2021training}. Therefore, there is a pressing need to develop more efficient transformers for inference and training.

\begin{figure}[ht]
	\centering
	\includegraphics[width=0.95\columnwidth]{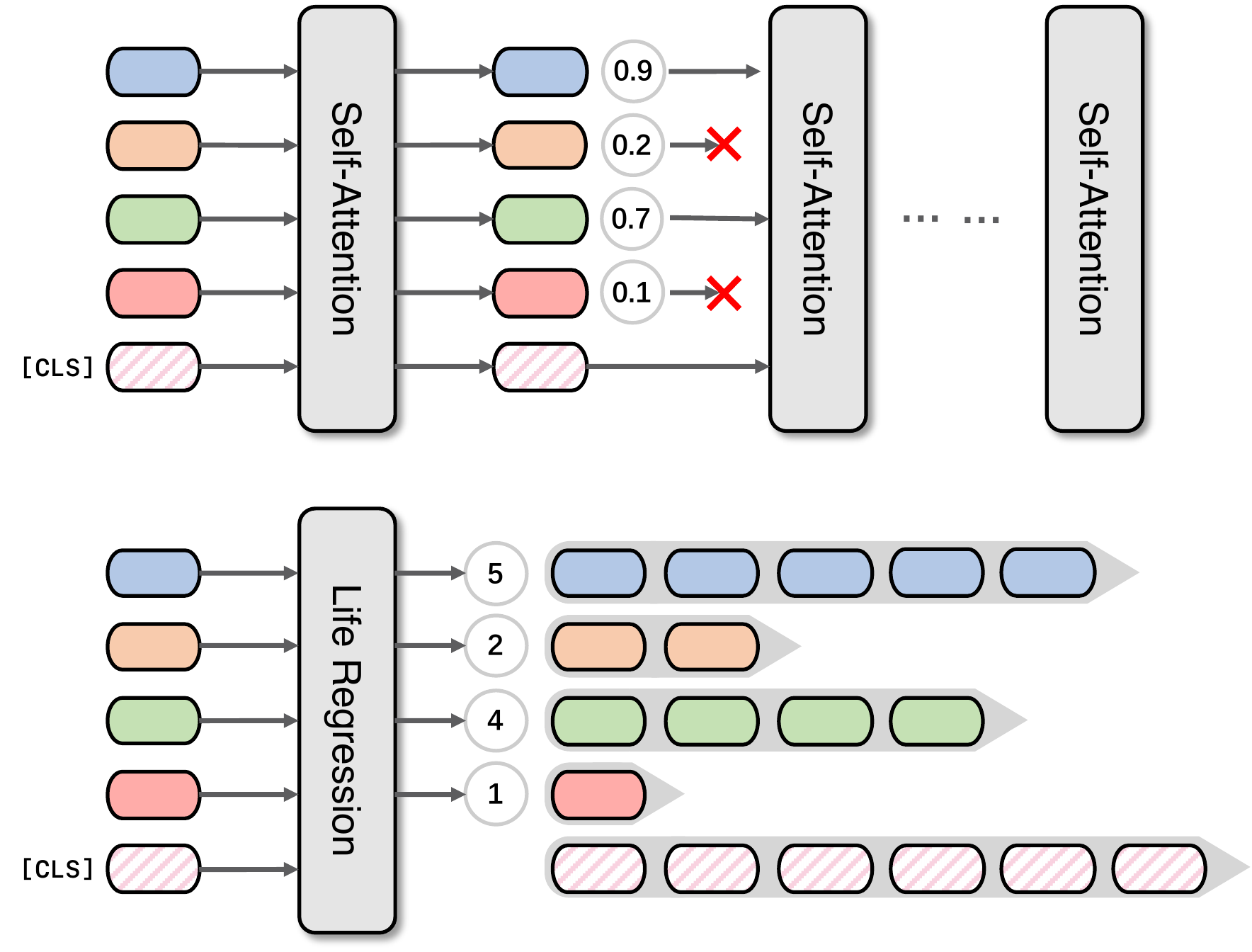}
	\caption{Comparison between the traditional framework and the proposed model. \textbf{Top}: the traditional score-per-layer model determines whether a specific patch should be discarded in the next layer by predicting a score for each patch. \textbf{Bottom}: the proposed life regression model predicts the life of all the patches in one go and discards patches according to their life.}
	\label{fig:architecture}
\end{figure}

Previous works~\cite{yu2022width,liang2022not,rao2021dynamicvit,zhu2021vision} have addressed the redundancy of input image patches in vision transformers and have attempted to optimize patch slimming to enhance computation efficiency. Most of these approaches~\cite{liang2022not,rao2021dynamicvit,zhu2021vision} have utilized the same paradigm, classifying image patches in each layer into two classes, one for keeping and another for discarding, using an attentive score. As illustrated in Figure \ref{fig:architecture} (top), these models retain patches with higher scores and discard those with lower scores.

However, this paradigm can result in $O(T'S)$ feature computation and slimming prediction time complexity, where $T'$ is the number of layers used for patch slimming, and $S$ is the slimming prediction time in each layer. The additional cost is proportional to the number of prediction layers $T'$ and creates noticeable overheads for accelerating vision transformers.

To address this problem and enhance inference speed further, we propose a new approach to patch slimming. We introduce a life regression module that predicts the life of individual patches in a single pass. As shown in Figure \ref{fig:architecture} (bottom), the proposed life regression module predicts the life of input patches, rather than determining whether to discard a specific patch layer-by-layer. The life of a patch reflects how long the patch can be utilized over layers. In the inference stage, once the current layer index exceeds a patch's predicted life, we discard the patch. In the training stage, we introduce a weight conversion module that converts a patch's predicted lifespan to an importance weight in each layer, enabling end-to-end differential optimization.

Our approach converts the traditional classification problem (whether to discard a patch in a specific layer) to a regression problem (how long will a patch be useful in the model?). In this way, our proposed model can avoid extra memory copying (gathering patches that survived each layer to new memory) and additional classification scores computation (projecting and computing features for each patch layer-by-layer). By effectively predicting patch lifetimes, our approach can reduce feature computation and the slimming prediction time complexity to $O(S)$. Through empirical study, we show that our model requires fewer fine-tuning epochs and achieves competitive performance with better efficiency.

Our paper presents new insights and solutions to the patch slimming problem that contributes to the advancement of vision transformer models. The contributions of this work can be summarized as follows:

\begin{itemize}
	\item We propose a new approach to the patch slimming problem by introducing a patch life regression model that converts the traditional patch classification problem into the patch life regression problem. This new perspective provides a fresh direction for future research beyond the traditional patch slimming paradigm.
	\item We develop a weight conversion module that converts patch lifetime predictions into patch importance weights for enabling end-to-end differential optimization. This module also resolves the non-differentiation problem of the life regression.
	\item Our proposed model offers competitive performance compared to other patch slimming models but with better inference efficiency and significantly fewer training epochs.
\end{itemize}

\section{Related Work}
\subsection{Vision Transformers}
Transformers were initially developed for natural language processing tasks in order to capture long-range dependencies between words~\cite{vaswani2017attention,devlin2018bert,brown2020language}. Since the introduction of Vision Transformer (ViT)~\cite{dosovitskiy2020image}, transformers have gained much attention in the computer vision community. Vision transformers split the input image into a grid of patches and feed them through multiple multi-head self-attention layers sequentially.

Considerable research has been carried out to improve vision transformers. For example, DeiT~\cite{touvron2021training} introduces distillation tokens which can be used to learn knowledge from teacher models. LV-ViT~\cite{jiang2021all} presents a novel training objective called token labeling, which exploits token-level recognition tasks to improve image classification performance. BoTNet~\cite{srinivas2021bottleneck} borrows the bottleneck structure from ResNet~\cite{he2016deep} and demonstrates that this structure can lead to better generalization empirically. CaiT~\cite{touvron2021going} introduces a class-attention layer which can separate the contradictory objectives of guiding attention. In addition, there are many works~\cite{lin2021eapt,jiao2023dilateformer,zhang2022tube,jiayao2022real,jia2022learning} focusing on the application of vision transformers.

\subsection{Patch Slimming}
Patch slimming is a popular research direction that aims to reduce computation and enhance inference speed by removing redundant image patches at suitable stages. 

Several recent works have attempted to address the patch slimming problem. The approaches proposed by \citeauthor{zhu2021vision} and \citeauthor{rao2021dynamicvit} predict the score for each patch feature in multiple layers and use an attention masking strategy to prune patches hierarchically. EviT~\cite{liang2022not} classifies patches into an attractive group and an unattractive group and fuses the features of the unattractive group while keeping the attractive group to improve efficiency and reduce performance degradation. Evo-ViT~\cite{xu2022evo} uses a slow-fast token evolution strategy to update the attractive and unattractive patch groups. A-ViT~\cite{yin2022vit} augments the vision transformer block with adaptive halting modules that compute a halting probability to prune unnecessary tokens. 

All of these methodologies have a similar framework: predicting which patch to discard in subsequent layers. Consequently, they cannot avoid additional patch prediction computations in the layers where patches are discarded.

\section{Methodology}
\begin{figure}[!tp]
	\centering
	\includegraphics[width=0.7\columnwidth]{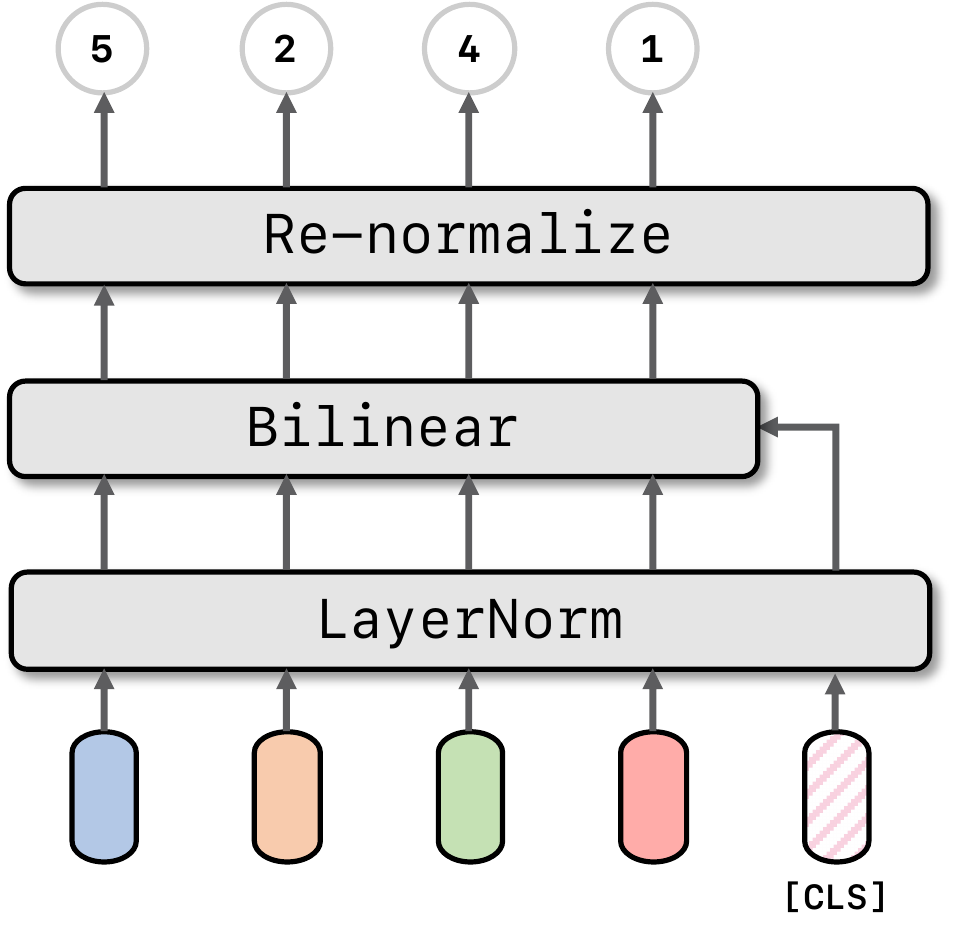}
	\caption{The proposed patch life regression module placed after layer $t_\text{base}$. All patches (except for the \lstinline{[CLS]} token) are sent to the module to compute their life. The life of the \lstinline{[CLS]} token is set to the maximum life $T$ manually.}
    \label{fig:model}
\end{figure}

Previous studies~\cite{liang2022not,rao2021dynamicvit} have shown that not all image patches contribute equally to the performance of vision transformers and certain patches can be redundant. By discarding less important patches at the appropriate stages, the inference speed of vision transformers can be significantly improved.

Most current patch slimming models use a layer-by-layer patch classification approach in which scores are predicted for each patch in each layer, and patches with low scores are discarded based on a predefined threshold. This classification paradigm is intuitive but requires feature transformation and score computation for every patch in each layer, leading to additional computational costs.

Our method is based on the hypothesis that early layer features of vision transformers can be used to predict the patch slimming scheme for higher layers in a single step. This approach allows us to avoid patch score prediction in every layer, reducing computational expenses and further accelerating inference speed. To achieve this, we propose to predict the lifespan of patches instead of their scores to perform patch slimming.

\subsection{Life Regression}
We denote the total number of patches from the input image as $N$ and use $t$ (the first letter of \emph{time}) to represent the layer index, as we describe our model forwarding as the life cycle of patches. The number of layers in the backbone vision Transformer is denoted by $T$. We use $i$ to represent the index of the $i$-th patch feature, where $i \in \left\{1,2,...,N \right\} \cup \left\{cls \right\}$, with $cls$ denoting the index of the \lstinline{[CLS]} token used for the final layer's image class prediction.

In our approach, we propose to predict the \textit{life} $\tau_i$ of each patch $i \in \left\{1,...,N\right\}$ in vision Transformers, while manually setting the life of \lstinline{[CLS]} to $T$. During inference, when the current layer index $t$ is greater than the life $\tau_i$, the patch $i$ is considered redundant and can be removed in the next layer. If the life $\tau_i$ is greater than or equal to $T$, the patch $i$ will be preserved throughout the model forwarding.

Our life regression module (illustrated in Figure \ref{fig:model}) is based on the assumption that a patch's life is positively correlated with its importance. To predict the life of a patch, we first predict its global importance score.

The layer index where our model predicts the life of patches in one go is denoted by $t_\text{base}$. We begin by normalizing the feature $x_i^{(t_\text{base})}$ in layer $t_\text{base}$ using layer normalization~\cite{ba2016layer}:

\begin{equation}
\bar{x}_i^{(t_\text{base})} = \text{LayerNorm}(x_i^{(t_\text{base})}).    
\end{equation}

To calculate the global importance score $s_i$ of a patch $i$, we first use a linear layer to project the patch features $\bar{x}_i^{(t_\text{base})}$, where $i \in \left\{1,2,...,N \right\} \cup \left\{cls \right\}$, to a semantic subspace. We consider the projected feature of the \lstinline{[CLS]} token as the representation of the full image. Next, we compute the inner product $s_i$ between the projected \lstinline{[CLS]} token and the projected image patch $i$, which indicates the similarity between the full image and patch $i$. This similarity value serves as the global importance score of patch $i$. 

To accomplish this, we can use a bilinear module, which is composed of two linear projection layers followed by an element-wise product. The output of this module is the similarity $s_i$ between the projected \lstinline{[CLS]} token and the projected image patch $i$: 

\begin{equation}
\label{eq:bilinear}
s_i = (\bar{x}_\text{cls}^{(t_\text{base})})^\intercal\cdot W \cdot \bar{x}_i^{(t_\text{base})}
, \\
i \in [1,...,N],
\end{equation}
where $s_i$ is the global importance score of the patch $i$, and $W$ is a trainable weight matrix. 

Directly using the global importance score $s_i$ of a patch $i$ as its life can lead to a problem. If the model predicts $s_i$ to be equal to the number of layers $T$, no patches will be discarded, and the model acceleration will be ineffective. To avoid this issue, we need to control the distribution of predicted patch life. 

To achieve this, we re-normalize the global importance score $s_i$ of each patch $i$, which produces a new importance score $\hat{s}_i$:

\begin{equation}
\label{eq:renormalize}
	\tau_i = \frac{s_i - \text{mean}(s)}{\text{std}(s)} \times \sigma + \mu,
\end{equation}
where $\sigma$ and $\mu$ are the target standard deviation and the target mean of all the predicted life $\left\{\tau_i\right\}_{i=1}^N$, which both can be viewed as hyper-parameters for now and will be explained in Section \ref{sec:inference}. 
In order to ensure that the model predicts longer lives for more important patches and shorter lives for redundant ones, we use $\text{mean}(s)$ and $\text{std}(s)$ to restrict the mean and standard deviation of the patch lives $\tau_i$, respectively.

During inference, the predicted life $\tau_i$ allows the model to discard patches whose expected contribution to the output has already been realized. However, in the training stage, directly discarding patches with short life can make the model non-differentiable and hard to optimize. To address this issue, we introduce a weight conversion mechanism to make the life prediction mechanism differentiable.

\subsection{Weight Conversion}

We define the function $\bar{\beta}_i(t)$ to indicate whether patch $i$ should be retained at layer $t$. Specifically, $\bar{\beta}_i(t) = 0$ denotes that patch $i$ is not used in layer $t$, while $\bar{\beta}_i(t) = 1$ denotes that patch $i$ is used in layer $t$:

\begin{equation}
\bar{\beta}_i(t) = 
\begin{cases}
    1 & t \le \tau_i, \\
	0 & t > \tau_i. \\
\end{cases}
\end{equation}

So if the current layer index $t$ is larger than the life $\tau_i$, the patch $i$ will no longer be used.

The self-attention module in layer $t$ of the vanilla vision transformer is 

\begin{equation}
\label{eq:before_plugin}
x^{(t+1)}_i = 
\sum_{j=1}^N 
\frac{\exp h^{(t)}_{ij}}{\sum_{k=1}^N \exp h^{(t)}_{ik}} 
\cdot 
x^{(t)}_j, \quad h^{(t)}_{ij} = (\kappa^{(t)}_i)^\intercal \cdot q^{(t)}_j.
\end{equation}
where $\kappa^{(t)}_i$ is the key feature of the patch $i$ in the layer $t$ and $q^{(t)}_j$ is the query feature of the patch $j$ in the layer $t$.

Consistent with \citeauthor{rao2021dynamicvit}, we can incorporate the function $\bar{\beta}_i(t)$ into the self-attention module to implement patch slimming without having to discard patches:

\begin{equation}
\label{eq:weighted_self_attention}
x^{(t+1)}_i = 
\sum_{j=1}^N 
\frac{\bar{\beta}_j(t) \cdot \exp h^{(t)}_{ij}}{\sum_{k=1}^N \bar{\beta}_k(t) \cdot  \exp h^{(t)}_{ik}} 
\cdot 
x^{(t)}_j.
\end{equation}

We can infer from Equation \ref{eq:weighted_self_attention} that setting $\bar{\beta}_j(t)$ to zero is equivalent to discarding patch $j$ in layer $t$, whereas setting $\bar{\beta}_j(t)$ to one is equivalent to keeping patch $j$ in layer $t$ by utilizing Equation \ref{eq:before_plugin}.

\begin{figure}[!tp]
	\centering
	\includegraphics[width=0.95\columnwidth]{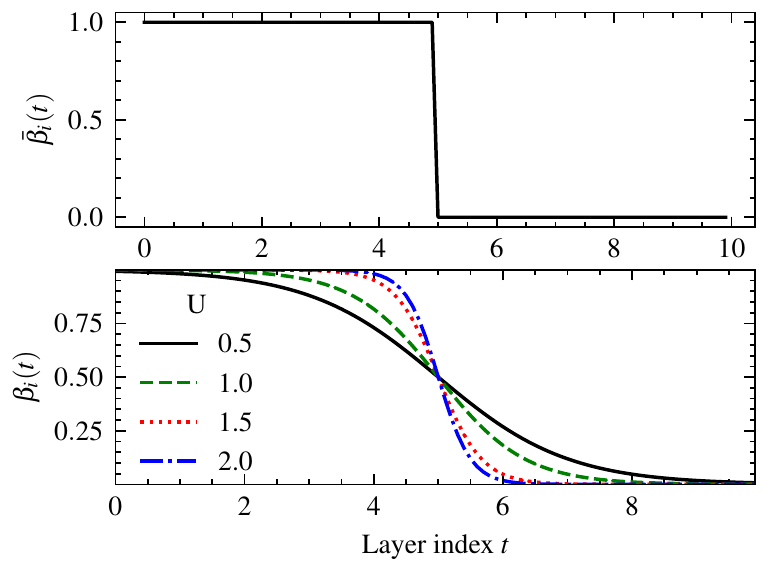}
	\caption{Function curve of the weight function $\bar{\beta}_i(t)$ and $\beta_i(t)$. \textbf{Top}: $\bar{\beta}_i(t)$ where $\tau_i = 5$; \textbf{Bottom}: $\beta_i(t)$ where $\tau_i = 5$ with different values of $U$.}
	\label{fig:beta}
\end{figure}

In Figure \ref{fig:beta} (top), we observe the function curve of $\bar{\beta_i}(t)$, which takes the shape of a horizontally-flipped step function. The breakpoint of this curve is equal to the life $\tau_i$, which is represented as 5 in the figure. In order to optimize the proposed model through end-to-end gradient descents, we need to devise a smoother function to approximate $\bar{\beta_i}(t)$.

Fortunately, the horizontally flipped and translated Sigmoid function $\beta_i(t)$ provides a reliable approximation for the function $\bar{\beta_i}(t)$:

\begin{equation}
\label{eq:beta}
\beta_i(t) = \frac{1}{1 + \exp(U(\tau_i - t))},
\end{equation}
where the temperature parameter $U$ is a hyper-parameter to adjust the steepness of $\beta_i(t)$ w.r.t. $t$. 
The curve of the function $\beta_i(t)$ with different values of $U$ is shown in Figure \ref{fig:beta} (bottom). 
From the figure, we can observe that $\beta_i(t)$ approaches $\bar{\beta_i}(t)$ as $U$ increases towards infinity. 
Therefore, when $U$ is large enough, $\beta_i(t)$ can serve as an excellent approximation for $\bar{\beta_i}(t)$. 
Based on this reason, we replace the function $\bar{\beta_i}(t)$ in Equation \ref{eq:weighted_self_attention} with the smooth approximation provided by the function $\beta_i(t)$.

\subsection{Training}

In our approach, we replace the function $\bar{\beta}_i(t)$ with the soft weight $\beta_i(t)$ for patch $i \in \left\{1,2,...,N \right\} \cup \left\{cls \right\}$ in each layer $t$, following Equation \ref{eq:weighted_self_attention}. 
Moreover, we do not discard patches during training. 
For training, our proposed model employs the cross-entropy loss, which is identical to the loss function used in the vanilla vision transformer:

\begin{equation}
\mathcal{L} = \text{CrossEntropy}(\hat{y}, y).
\end{equation}

To avoid interference between the proposed module and the backbone transformer, we employ a two-stage training scheme as detailed in Algorithm \ref{alg:algorithm}. 
During the first stage, we only fine-tune the life predictor $P$, which comprises the life regression module and weight conversion module, while the parameters of the backbone vision transformer $M$ are frozen. 
In the second stage, we fine-tune the vision transformer $M$ while retaining the life predictor $P$ parameters as fixed.

\begin{algorithm}[tb]
\caption{Two-stage Training}
\label{alg:algorithm}
\textbf{Input}: Dataset $\mathcal{D}$, pre-trained model $M$, the life predictor $P$. \\
\textbf{Output}: Fine-tuned model $M$, trained $P$. \\
\begin{algorithmic}[1] %[1] enables line numbers
	\STATE Load pre-trained weights to $M$ and randomly initialize parameters of $P$.
	\STATE Freeze all parameters in $M$.
	\STATE Train $P$ by gradient descents until convergence.
	\STATE Unfreeze all parameters in $M$ and freeze all parameters in $P$.
	\STATE Fine-tune $M$ until convergence.
\end{algorithmic}
\end{algorithm}

\subsection{Inference}\label{sec:inference}
During inference, we may need to discard certain patches to accelerate the vision transformer. However, as the different input images could have different distributions of life prediction, the same layer may require different patches to be discarded for different samples, rendering batch inference impractical.

To facilitate batch inference, we constrain the number of patches used in layer $t$ to a fixed value $n_t$, which is consistent across all input images, where $t = 1,...,T$. Moreover, we ensure that $[n_t]_{t=1}^T$ satisfies $n_{t_1} \ge n_{t_2}$ for $1 \le t_1 < t_2 \le T$.
With the fixed number of patches $[n_t]_{t=1}^T$, the mean $\mu$ and the standard deviation $\sigma$ of the life predictions used in Equation \ref{eq:renormalize} can be computed by:

\begin{equation}
    \mu = \frac{1}{N}\left( \sum_{t=1}^{T-1} t \cdot (n_t - n_{t+1}) + T \cdot n_T \right)
\end{equation}
and 

\begin{equation}
    \sigma = \sqrt{ \frac{1}{N}\left( \sum_{t=1}^{T-1} (t - \mu) \cdot (n_t - n_{t+1}) + (T - \mu) \cdot n_T \right) }
\end{equation}

In order to compare with baseline models, patches with the same keep rate $\rho$ ($0 < \rho \le 1$) are discarded at layers $[t_i]_{i=1}^{T'} = [t_1,...,t_{T'}]$, where $n_{t_{i+1}} = \rho n_{t_{i}}$ and $n_{t'} = n_{t_{i}}$ for all $t_{i} \le t' < t_{i+1}$. Only patches with a top-$n_{t}$ weight $\beta_i(t)$ in layer $t$ are retained.

As we manually control the number of patches in each layer, we need to compute the patch weight $\beta_i(t)$ for patch $i$ in layer $t$ and only keep those with larger $\beta_i(t)$. However, directly computing $\beta_i(t)$ can be time-consuming during inference. 

\begin{table*}[tb]
    \centering
    \caption{Comparison between the proposed model and baselines with DeiT-\{S, B\}. The `-' symbol indicates that the field is not applicable for this model or the corresponding data is not available for this model. Each section contains the models with the same backbone (the rows of backbones are highlighted with \colorbox{lightgray!20}{gray background}). `Epochs’ denotes the number of training epochs. Each epoch costs about 1$\sim$2 hours.}
    \label{tab:sota_deit}
    \begin{NiceTabular}{lccccc}
        % \CodeBefore
        % \rowcolor{lightgray!20}{2,7}
        % \Body
        \toprule
        \textbf{Method} & \textbf{\makecell{Top-1 Acc (\%) $\uparrow$}} & \textbf{Top-5 Acc (\%) $\uparrow$} & \textbf{FLOPs (G)} $\downarrow$ & \textbf{\makecell{Throughput (img/s) $\uparrow$}} & \textbf{\makecell{Epochs $\downarrow$}}  \\
        \midrule
        \cellcolor{lightgray!30}DeiT-S & \cellcolor{lightgray!30}79.8 & \cellcolor{lightgray!30}95.0 & \cellcolor{lightgray!30}4.6 & \cellcolor{lightgray!30}503.9 & \cellcolor{lightgray!30}- \\
        IA-RED\textsuperscript{2} (DeiT-S) & 79.1 \blue{(-0.7)} & 94.5 \blue{(-0.5)} & 3.3 \blue{(-39\%)} & - & 30 \\
        EViT (DeiT-S) & 79.2 \blue{(-0.6)} & 94.7 \blue{(-0.3)} & 3.0 \blue{(-53\%)} & 658.2 \blue{(+31\%)} & 30 \\
        DynamicViT (DeiT-S) & 79.3 \blue{(-0.5)} & 94.6 \blue{(-0.4)} & 3.4 \blue{(-35\%)} & 642.4 \blue{(+27\%)} & 30 \\
        Ours (DeiT-S) & 79.3 \blue{(-0.5)} & 94.7 \blue{(-0.3)} & 3.2 \blue{(-44\%)} & 704.8 \blue{(+40\%)} & 2 \\
        \midrule
        \cellcolor{lightgray!30}DeiT-B & \cellcolor{lightgray!30}82.0 & \cellcolor{lightgray!30}95.7 & \cellcolor{lightgray!30}17.6 & \cellcolor{lightgray!30}186.4 & \cellcolor{lightgray!30}- \\
        IA-RED\textsuperscript{2} (DeiT-B) & 80.3 \blue{(-1.7)} & 95.0 \blue{(-0.7)} & 11.8 \blue{(-49\%)} & - & 30 \\
        Evo-ViT (DeiT-B) & 80.0 \blue{(-2.0)} & 94.7 \blue{(-1.0)} & 11.7 \blue{(-50\%)} & 273.1 \blue{(+47\%)} & 30 \\
        Ours (DeiT-B) & 80.1 \blue{(-1.9)} & 94.7 \blue{(-1.0)} & 12.3 \blue{(-43\%)} & 262.0 \blue{(+41\%)} & 2 \\
        \bottomrule
    \end{NiceTabular}
\end{table*}

We use the following property to avoid computing $\beta_i(t)$ and only compute the life $\tau_i$. 
\begin{proposition}
	$\tau_i \ge \tau_j$ $\Leftrightarrow$ for all $t \in \mathbb{R}$, $\beta_i(t) \ge \beta_j(t)$.
\end{proposition}
The property suggests that when comparing the weights $\beta_i(t)$ and $\beta_j(t)$ for patch $i$ and $j$, we only need to compare their respective lives $\tau_i$ and $\tau_j$.

Therefore, during inference, to determine which patch to discard, we only need to compute and compare the patch's life $\tau_i (i=1,...,N)$, and we do not need to compute the weight $\beta_i(t)$, which is only calculated during training.

\section{Experiment}
In our experiment, we perform comparative analyses with various vision transformer backbones, all under the same experimental setting. In addition, we conduct ablation studies to investigate the effects of hyper-parameters $\rho$, $U$, and different training schemes. Our experiments are carried out using the widely used ImageNet~\cite{krizhevsky2017imagenet} dataset.

\subsection{Implementation Detail}
Our default values for $U$ and $[t_i]_{i=1}^{T'}$ are set to 1.5 and $[4, 7, 10]$, respectively. We employ the AdamW~\cite{loshchilov2017decoupled} optimizer, with a learning rate of $1\mathrm{e}{-5}$ for both the life regression module and weight conversion module. The transformer backbone is trained with a learning rate of $1\mathrm{e}{-3}$. Our models are trained and evaluated with a batch size of 128 on an Nvidia Quadro P6000 GPU.

\begin{table*}[tb]
    \centering
    \caption{Comparison between the proposed model and baselines with ViT-\{T, S, B\}. The `-' symbol indicates that the field is not applicable for this model or the corresponding data is not available for this model. Each section contains the models with the same backbone (the rows of backbones are highlighted with \colorbox{lightgray!20}{gray background}). `Epochs’ denotes the number of training epochs. Each epoch costs about 1$\sim$2 hours.}
    \label{tab:sota_vit}
    \begin{NiceTabular}{lccccc}
        % \CodeBefore
        % \rowcolor{lightgray!20}{2,7,11}
        % \Body
        \toprule
        \textbf{Method} & \textbf{\makecell{Top-1 Acc (\%) $\uparrow$}} & \textbf{Top-5 Acc (\%) $\uparrow$} & \textbf{FLOPs (G)} $\downarrow$ & \textbf{\makecell{Throughput (img/s) $\uparrow$}} & \textbf{\makecell{Epochs $\downarrow$}}  \\
        \midrule
        \cellcolor{lightgray!30}ViT-T & \cellcolor{lightgray!30}75.5 & \cellcolor{lightgray!30}92.8 & \cellcolor{lightgray!30}1.3 & \cellcolor{lightgray!30}1212.1 & \cellcolor{lightgray!30}- \\
        DynamicViT (ViT-T) & 70.9 \blue{(-4.6)} & 90.0 \blue{(-2.8)} & 0.9 \blue{(-44\%)} & 1533.9 \blue{(+27\%)} & 30 \\
        A-ViT (ViT-T) & 71.4 \blue{(-4.1)} & 90.4 \blue{(-2.4)} & 0.8 \blue{(-63\%)} & - & 30 \\
        Evo-ViT (ViT-T) & 72.0 \blue{(-3.5)} & 90.7 \blue{(-2.1)} & 0.8 \blue{(-63\%)} & 1334.1 \blue{(+10\%)} & 30 \\
        Ours (ViT-T) & 73.6 \blue{(-1.9)} & 91.8 \blue{(-1.0)} & 0.9 \blue{(-44\%)} & 1573.8 \blue{(+30\%)} & 2 \\
        \midrule
        \cellcolor{lightgray!30}ViT-S & \cellcolor{lightgray!30}81.4 & \cellcolor{lightgray!30}96.1 & \cellcolor{lightgray!30}4.6 & \cellcolor{lightgray!30}504.7 & \cellcolor{lightgray!30}- \\
        Evo-ViT (ViT-S) & 79.5 \blue{(-1.9)} & 94.8 \blue{(-1.3)} & 3.0 \blue{(-53\%)} & 733.4 \blue{(+45\%)} & 30 \\
        A-ViT (ViT-S) & 78.8 \blue{(-2.6)} & 93.9 \blue{(-2.2)} & 3.6 \blue{(-28\%)} & - & 30 \\
        Ours (ViT-S) & 79.8 \blue{(-1.6)} & 95.3 \blue{(-0.8)} & 3.2 \blue{(-44\%)} & 723.3 \blue{(+43\%)} & 2 \\
        \midrule
        \cellcolor{lightgray!30}ViT-B & \cellcolor{lightgray!30}84.5 & \cellcolor{lightgray!30}97.3 & \cellcolor{lightgray!30}17.6 & \cellcolor{lightgray!30}187.1 & \cellcolor{lightgray!30}- \\
        Evo-ViT (ViT-B) & 81.6 \blue{(-2.9)} & 95.6 \blue{(-1.7)} & 12.0 \blue{(-47\%)} & 219.0 \blue{(+17\%)} & 30 \\
        EViT (ViT-B) & 82.4 \blue{(-2.1)} & 96.0 \blue{(-1.3)} & 12.1 \blue{(-45\%)} & 248.2 \blue{(+33\%)} & 30 \\
        Ours (ViT-B) & 82.3 \blue{(-2.2)} & 96.4 \blue{(-0.9)} & 12.3 \blue{(-43\%)} & 256.0 \blue{(+37\%)} & 2 \\
        \bottomrule
    \end{NiceTabular}
\end{table*}

\subsection{Comparison with SOTA methods}
In order to demonstrate the effectiveness and efficiency of our proposed model, we conduct comparisons with various vision transformer backbones and state-of-the-art (SOTA) models under the same experimental conditions. The models we compare against are:

\begin{itemize}
    \item ViT-\{T, S, B\}~\cite{dosovitskiy2020image}: ViT is the pioneering vision transformer that established the viability of adapting the transformer architecture, originally proposed for NLP~\cite{vaswani2017attention}, for the computer vision domain with competitive performance.
    \item DeiT-\{S, B\}~\cite{touvron2021training}: DeiT introduced a teacher-student distillation strategy to vision transformers that significantly reduced training time while maintaining comparable performance.
    \item DynamicViT~\cite{rao2021dynamicvit}: DynamicViT presents a novel hierarchical token sparsification strategy that dynamically removes superfluous tokens.
    \item IA-RED\textsuperscript{2}~\cite{pan2021ia}: IA-RED\textsuperscript{2} empirically demonstrates the ability of the patch slimming approach to produce interpretable patterns.
    \item A-ViT~\cite{yin2022vit}: A-ViT uses an adaptive token discarding framework to expedite inference in vision transformers.
    \item Evo-ViT~\cite{xu2022evo}: Evo-ViT performs token pruning through a slow-fast token evolution methodology.
    \item EViT~\cite{liang2022not}: EViT exploits the attention map generated by vision transformers to determine patch slimming configurations in each layer.
\end{itemize}

Table \ref{tab:sota_deit} reports the evaluation results for models based on the DeiT-S and DeiT-B backbones, while Table \ref{tab:sota_vit} presents results for models built on the ViT-T, ViT-S, and ViT-B backbones, respectively. In the first section of Table \ref{tab:sota_deit}, we observe that our proposed model achieves a Top-1 accuracy of 79.3\% and Top-5 accuracy of 94.7\%, which are among the best results when compared to all other methods. With only a 0.5\% drop in Top-1 accuracy, our approach achieves a 40\% speedup measured by throughput, surpassing the performance of other patch slimming methods. Notably, our approach requires only two epochs (approximately 4 hours) for finetuning, while other methods typically need around 30 epochs (roughly 60 hours) of training time.

Moving to the second section of Table \ref{tab:sota_deit}, which considers models based on DeiT-B, our proposed model achieves a competitive result with a Top-1 accuracy of 80.1\% and Top-5 accuracy of 94.7\%. While the throughput of our model is slightly slower than other methods, it still obtains a 41\% throughput speedup and only requires two epochs to finetune.

Regarding the ViT-T-based models, as shown in the first section of Table \ref{tab:sota_vit}, our proposed model demonstrates superior performance compared to the baselines, achieving a 1.6$\sim$2.7\% increase in Top-1 accuracy while exhibiting higher throughput than DynamicViT and Evo-ViT. The second and third sections of Table \ref{tab:sota_vit} illustrate results for the proposed model with ViT-S and ViT-B backbones, respectively, showing that our approach achieves significant inference acceleration without significantly sacrificing performance, requiring only two epochs to finetune. Compared to A-ViT, our proposed model achieves both higher accuracy and lower FLOPS, while outperforming Evo-ViT in terms of performance and exhibiting competitive throughput.

Overall, our proposed model achieves competitive or superior performance compared to baselines with different backbones, achieving significant inference acceleration with little compromise in performance. Moreover, our approach only requires two epochs to finetune, which is much fewer than the number of epochs required by the baselines.

\begin{figure*}[tb]
	\centering
	\includegraphics[width=0.8\textwidth]{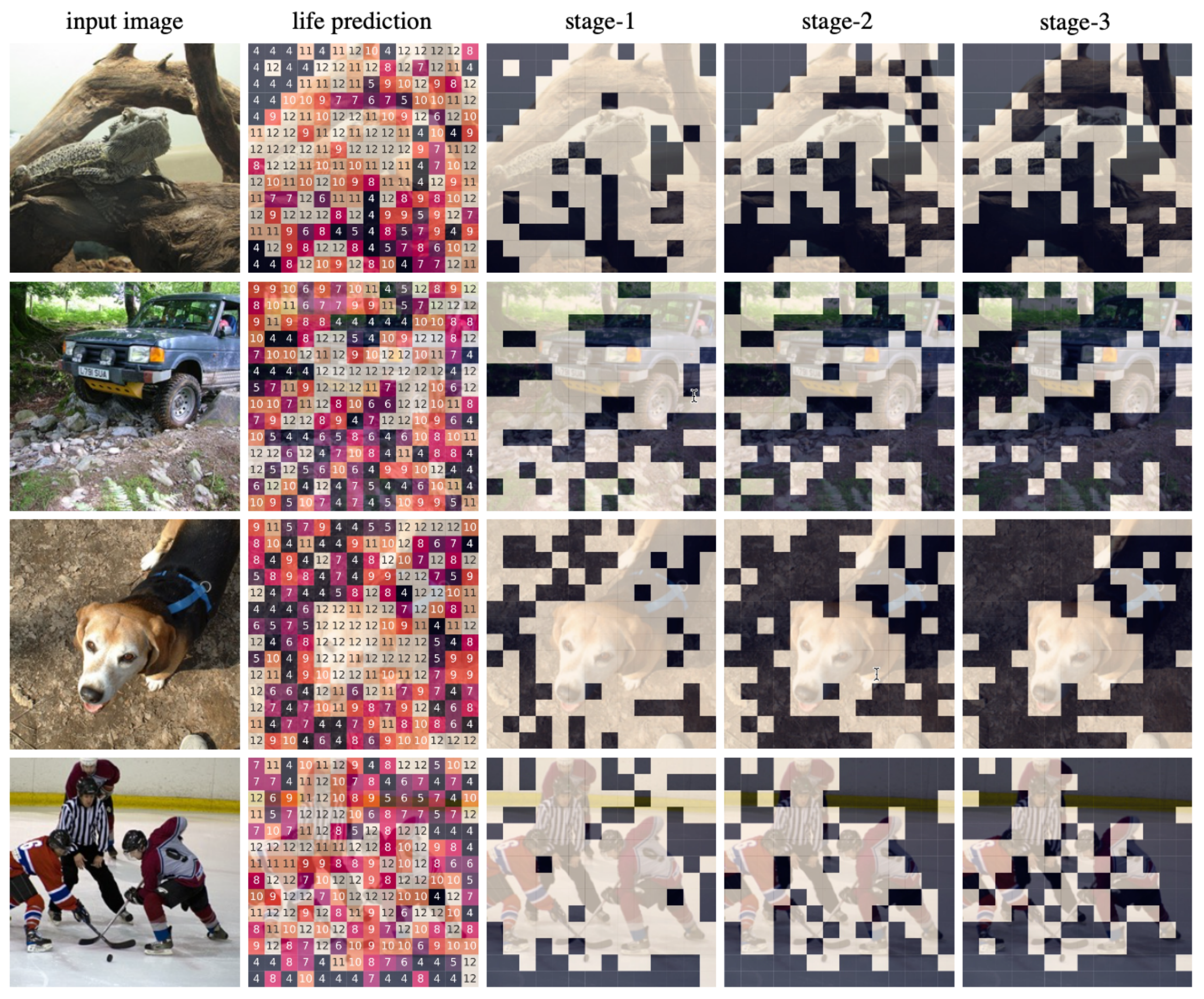}
	\caption{Visualization of the lives predicted by the proposed model with the DeiT-S backbone. The numbers in the cell of the life prediction images are the $\tau_i$ predicted by the proposed model. Stage-\{1,2,3\} represent the patch slimming stages in layer \{4, 7, 10\}. The light cell indicates the patch is kept in this stage, and the dark cell indicates the patch is discarded in the stage.}
	\label{fig:visualization}
\end{figure*}

\subsection{Ablation Study}
We perform the ablation study on the keep rate $\rho$, the temperature parameter $U$ and the training scheme as follows.

\subsubsection{Keep rate}
\begin{table}[!t]
    \centering
    \caption{The proposed model is set to different keep rate $\rho$, and the backbone is DeiT-S. $\rho = 1.0$ indicates the original DeiT-S without patch slimming. }
    \label{tab:keep_rate}
    \begin{tabular}{lrrrr}
        \toprule
        \textbf{Keep rate $\rho$} & 1.0 & 0.9 & 0.8 & 0.7 \\
        \midrule
        \textbf{Top-1 Acc $\uparrow$} & 79.8 & 79.7 & 79.5 & 79.3 \\
        \textbf{Top-5 Acc $\uparrow$} & 95.0 & 94.8 & 94.8 & 94.7 \\
        \textbf{FLOPs (G) $\downarrow$} & 4.6 & 4.1 & 3.6 & 3.2 \\
        \textbf{\makecell{Throughput (img/s) $\uparrow$}} & 503.9 & 512.5 & 601.3 & 704.8 \\
        \bottomrule
    \end{tabular}
\end{table}

Table \ref{tab:keep_rate} displays the impact of the keep rate $\rho$ on performance and inference speed. We fix the DeiT-S backbone and vary the keep rate. As shown in the table, decreasing $\rho$ results in a marginal drop in performance but a significant increase in inference speed. This is because a lower keep rate leads to fewer patches to compute during inference, resulting in a faster computation time.

\subsubsection{Training Schemes}
We conducted an ablation study to evaluate the effectiveness of the two-stage training scheme outlined in Algorithm \ref{alg:algorithm}, and the results are presented in Table \ref{tab:training_scheme}. The table clearly indicates that the two-stage training approach consistently outperforms the simple single-stage methods, regardless of the keep rate. This phenomenon can be attributed to the fact that in the single-stage training strategy, the training of the life prediction module and the finetuning of the backbone module are coupled together, which can lead to interference between the two modules if they are trained simultaneously. In contrast, the two-stage training scheme decouples the training of the two modules and thus avoids such interference, leading to better performance.

\begin{table}[!t]
	\centering
	\caption{Abalation study of the training scheme. ``Single Stage" indicates that we finetune the life prediction module and the backbone simultaneously. ``Two Stages" indicates the strategy described in Algorithm \ref{alg:algorithm}.}
	\label{tab:training_scheme}
	\begin{tabular}{lrrrrrr}
		\toprule
		 & \multicolumn{3}{c}{\textbf{Single Stage}} & \multicolumn{3}{c}{\textbf{Two Stages}} \\
		 \midrule
		\textbf{Keep rate $\rho$} & 0.9 & 0.8 & 0.7 & 0.9 & 0.8 & 0.7 \\
%		\midrule
		\textbf{Top-1 Acc $\uparrow$} & 78.7 & 78.5 & 78.3 & 79.8 & 79.5 & 79.3 \\
		\textbf{Top-5 Acc $\uparrow$} & 93.4 & 92.1 & 91.9 & 95.2 & 95.0 & 94.7 \\
		\bottomrule
	\end{tabular}
\end{table}

\subsubsection{Temperature Parameter}
We conducted an ablation study to examine the impact of the temperature parameter $U$ on the proposed model. The parameter $U$ controls how accurately the function $\beta_i(t)$ approximates the function $\bar{\beta}_i(t)$, as illustrated in Figure \ref{fig:beta}. As $U$ increases, the approximation becomes more accurate, but the resulting function curve becomes flatter on the two sides of the curve of $\beta_i(t)$, which can make the optimization more challenging~\cite{han1995influence}. Table \ref{tab:U_acc} summarizes the results of our experiment with different $U$s. We observed that as $U$ increases, the performance of the proposed model first improves and then starts to decrease. In the stage of performance improvement, larger $U$ leads to better function approximation, and the flat curve effect does not come into play. However, in the stage of performance decline, the flat curve effect dominates the model performance.

\begin{table}[!htbp]
	\centering
	\caption{Experiment results of different temperature parameters $U$. The backbone is DeiT-S.}
	\label{tab:U_acc}
	\begin{tabular}{lrrrrr}
		\toprule
		\textbf{Temperature U} & 0.5 & 1 & 1.5 & 2 & 3 \\ 
		\midrule
		\textbf{Top-1 Acc} & 79.0 & 79.2 & \textbf{79.3} & 79.2 & 79.2 \\
		\textbf{Top-5 Acc} & 94.4 & 94.6 & \textbf{94.7} & 94.5 & 94.5 \\
		\bottomrule
	\end{tabular}
\end{table}

\subsection{Visualization}
Figure \ref{fig:visualization} provides a visual representation of our proposed method, showing four sample images with their corresponding life predictions for patches and the patch slimming results. The numbers inside the cells of life prediction images represent the predicted life for the patches. We clamp the life prediction to the range of 4 to 12, as predicted at layer 4 with 12 layers.

The life prediction figures demonstrate that our proposed model predicts longer life for patches that are more important for image classification. Specifically, the lizard in the first row, the Jeep in the second row, the dog in the third row, and the hockey players in the last row are predicted with longer life.

The stage-\{1, 2, 3\} figures illustrate the patch slimming results in each stage, where we discard the patch in layers 4, 7, and 10 according to the patch's predicted life. In the first row, the lizard and the wood are kept in all three stages. In the second row, the central part of the Jeep is kept. In the third row, the head of the Labrador is predicted to be important and retained. In the last row, the proposed model focuses on the hockey players and decides to keep the patches containing them. This highlights that our proposed model is able to focus on the correct concepts in images for classification purposes. 

\section{Conclusion}
In this work, we introduce a new paradigm to the vision transformer patch slimming task by transforming the traditional layer-by-layer patch classification task to the life regression task. Our proposed approach includes a life regression module used to predict the life of image patches and a weight conversion module designed to enable end-to-end gradient descent optimization. The experimental evaluations show that our approach accelerates the model inference with minimal performance compromise and requires fewer fine-tuning epochs than the baselines. We believe our work offers a promising direction towards accelerated vision transformer models.

\bibliographystyle{IEEEtranN}
\bibliography{ref}
\end{document}